\documentclass[conference]{IEEEtran}
\IEEEoverridecommandlockouts
\usepackage{cite}
\usepackage{amsmath,amssymb,amsfonts}
\usepackage{graphicx}
\usepackage{xcolor}
\usepackage{comment, lipsum}
\usepackage{amsmath,amssymb} 
\usepackage{tikz}
\usepackage{color}
\usepackage{multirow, multicol}
\usepackage{graphicx, caption, subcaption}
\usepackage{newtxtext,newtxmath}
\usepackage{subcaption}
\usepackage{newtxmath}
\usepackage{algorithm}
\usepackage{algpseudocode}
\usepackage{varwidth}
\usepackage{tabulary,booktabs}

\begin{document}

\title{Remote PhotoPlethysmography from Low Resolution videos: An end to end solution using Efficient ConvNets
}

\author{
\IEEEauthorblockN{Ruijia Deng}
\IEEEauthorblockA{
\textit{Wuhan University} \\
Wuhan, China}
\and
\IEEEauthorblockN{Bharath Ramakrishnan}
\IEEEauthorblockA{
\textit{Texas A\&M University}\\
Texas, USA}
}

\maketitle

\begin{abstract}
Measurement of the cardiac pulse from facial video has become an interesting pursuit of research over the last few years. This is mainly due to the increasing importance of obtaining the heart Rate of an individual in a non-invasive manner, which can be highly useful for applications in gaming and the medical industry. Another instrumental area of research over the past few years has been the advent of Deep Learning and using Deep Neural networks to enhance task performance. In this work, we propose to use efficient convolutional networks to accurately measure the heart rate of user from low resolution facial videos. Furthermore, to ensure that we are able to obtain the heart rate in real time, we compress the deep learning model by pruning it, thereby reducing its memory footprint. We benchmark the performance of our approach on the MAHNOB dataset and compare its performance across multiple approaches. 
\end{abstract}

\section{Introduction}

The heart rate of an individual is one of the most important physiological parameters to determine the health of the individual. The heart rate of an individual gives information about their current physical state as well as the overall fitness of the individual. For example, someone who engages in regular physical activity such as athletics will have a lower rest heart rate as compared to an individual who generally does not. In addition to determining the general health of a person, measures such as Heart Rate variability have been instrumental in indicating the mental makeup of a person, signifying the advent of depression, emotion changes, etc. as well as breathing disorders such as sleep apnea. \cite{icaart} employs a visual heart rate based system to detect changes in the user's heart rate variance during gameplay, which can then be used to design video games for optimal user experience. The aforementioned applications require long term monitoring due to which it can become inconvenient for the user to have heart rate sensors attached to them for extended periods of time. \cite{balakrishnan2013detecting, cennini2010heart, subramaniam2019spectral}.  
Two popular approaches to measure heart rate from facial video (remote photoplethysmography as its called) have been based on getting the HR from microscopic head motions of the user and heart rate retrieval from microscopic changes in the color of the user's face in response to the pulse of the blood flowing through their face. The above two approaches have been described below: 
There have been several approaches to measuring Heart Rate from an RGB video of the user's face. The two main approaches have been the following:
\begin{itemize}
    \item \textbf{Motion-based Estimation}: Inspired from \cite{balakrishnan2013detecting}, several approaches have documented the use of head pulse motions to measure heart rate. The primary concept of the approach lies in the fact that the head asserts a Newtonian reaction to the pulse of the blood flowing through the user's face as a result of the heart rate. The rate at which the head oscillates is related to the force/rate at which the heart pumps blood through the face. This method, while technically sound, can be problematic in the presence of motion artifacts since it would be difficult to observe the microscopic motions of the user's head due to their heart beat. 
    \item \textbf{Color-based Estimation}: Recently, a majority of approaches have used the remote photoplethysmographic (rPPG) signals from the RGB pixel intensity of the user's face. Its been found that the rPPG signal corresponding to the green channel of the facial video provides the most relevant information about the user's facial video \cite{lam2015robust, subramaniam2019spectral}. While a number of approaches have accurately measured the facial video, obtaining accurate HR measurements in the presence of significant motion and illumination interferences has been an issue.   
\end{itemize}

\begin{figure*}[h]
    \centering
    \includegraphics[scale=0.5]{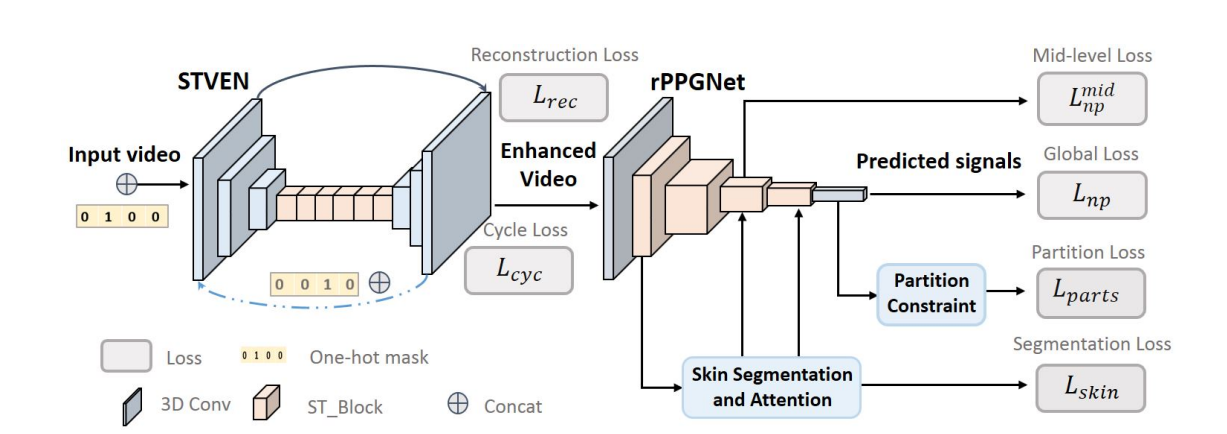}
    \caption{Architecture of the spatio-temporal model required for enhancing and reconstructing the quality of the input RGB video. The design of the model has been adopted from \cite{yu2019remote}}
    \label{fig:sppg}
\end{figure*}

With the advent of deep learning, there have been several approaches which seek to optimize system performance based on hardware \cite{raj2019programming} and software \cite{subramaniam2017neuromorphic, lee2018snip}, not just in the area of computer vision, but in other areas as well. In the proposed method, we follow a software-based approach to achieving optimal performance in remote photoplethysmography. We employ video-to-video generator networks called Spatio-Temporal Video Enhancement Networks from \cite{yu2019remote}. These networks are pruned using skip connection based pruning in \cite{subramaniam2020n2nskip}.

In this work, we propose to use efficient convolutional networks
to accurately measure the heart rate of user from low resolution
facial videos. Furthermore, to ensure that we are able to obtain
the heart rate in real time, we compress the deep learning
model by pruning it, thereby reducing its memory footprint. We
benchmark the performance of our approach on the MAHNOB \cite{soleymani2011multimodal}
dataset and compare its performance across multiple approaches. Our contributions are two-fold.

\begin{itemize}
    \item We propose an approach to measure the heart rate of a user from their facial video using an end-to-end deep learning solution
    \item We use a skip connection based method to compress the deep neural network in order to reduce its memory footprint. This ensures that the latency of the method is reduced so that the heart rate of the user can be measured in real time.
\end{itemize}

\section{Method}
We use a video generation neural network inspired from \cite{yu2019remote}. The structure of this neural network is shown in \ref{fig:sppg}. There are two models in the proposed method: one is an enhancement model used to enhance the quality of the video and the other model is a recovery model. In order to improve the recovery performance of the second model, we perform joint training on both the models. The joint loss consists of a reconstruction loss, which is a root mean square loss (RMSE) and a loop loss for improving the quality of the reconstruction.

\section{Experimental Results}

\subsection{Experimental Setup}
We train the spatio-temporal deep neural network by first pruning the model based on the approach of \cite{subramaniam2020n2nskip}. We add sparse convolutional connections to the model and prune parameters based on the importance of the connection sensitivity of each weight. We apply a sparsity/compression ratio of 10x so that we preserve 1 among 10 connections in the model. In other words, we reduce the memory footprint of the model by 10$\times$.  

\subsection{Results and Analysis}

Table \ref{tab:mahnob} shows the results of our dataset as compared to previous methods.

\setlength{\tabcolsep}{3.5pt}
\begin{table}[htbp]
\begin{center}
\begin{tabular}{ccccc}
\hline
Framework & $\mu_{error}$ & RMSE (\%) & \% Absolute Error<5bpm & r \\ 
\hline
\vspace{1mm}
Li2014 & 7.7 & 15 & 68.1 & $0.72^{*}$\\
\vspace{1mm}
SAM2016 & 5.5 & 14.2 & 72.1 & $0.75^{*}$\\
\vspace{1mm}
Zha2017 & 5.6 & 14.2 & 78.1 & $0.79^{*}$\\
\vspace{1mm}
Lam2015 & 6.7 & 14.1 & 63.1 & $0.75^{*}$\\
\vspace{1mm}
Sub2019 & 4.7 & 7.8 & 88.1 & $0.82^{*}$\\
\vspace{1mm}
Ours & 3.9 & 6.5 & 93.7 & $0.85^{*}$\\
\hline
\end{tabular}
\end{center}
\caption{Results of measuring HR from facial video on the MAHNOB-HCI dataset.}
\label{tab:mahnob}
\end{table}

\begin{figure}[htp]
    \centering
    \includegraphics[scale=0.5]{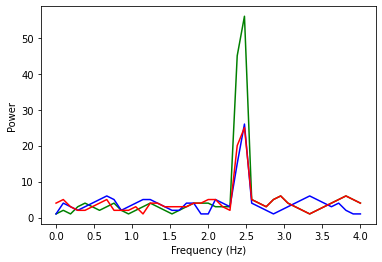}
    \caption{Plot of the heart rate of the user as a power spectra of the rPPG signals corresponding to each RGB channel.}
    \label{fig:HR}
\end{figure}

\section{Conclusion}
In this work, we propose to use efficient convolutional networks
to accurately measure the heart rate of user from low resolution
facial videos. Furthermore, to ensure that we are able to obtain
the heart rate in real time, we compress the deep learning
model by pruning it, thereby reducing its memory footprint. We
benchmarked the performance of our approach on the MAHNOB-HCI dataset and were able to outperform previous state-of-the-art approaches. Furthermore, we employed a single-shot pruning method to reduce the memory footprint of the spatio-temporal model so that we could reduce the memory of the model. As a result, the storage and latency of the framework was reduced greatly.

\bibliography{aaai2}


\end{document}